\documentclass{article}
\usepackage{spconf,amsmath,graphicx}

\title{DOVER: A Method for Combining Diarization Outputs}
\name{Andreas Stolcke and Takuya Yoshioka}

\address{Speech and Dialog Research Group\\
Microsoft\\
{\small\tt \{anstolck,tayoshio\}@microsoft.com}}

\begin{document}
\maketitle
\begin{abstract}
Speech recognition and other natural language tasks have long benefited
from voting-based algorithms as a method to aggregate outputs from
several systems to achieve a higher accuracy than any of the individual
systems. Diarization, the task of segmenting an audio stream into
speaker-homogeneous and co-indexed regions, has so far not seen the benefit
of this strategy because the structure of the task does not lend itself
to a simple voting approach. This paper presents DOVER (diarization
output voting error reduction), an algorithm for weighted voting
among diarization hypotheses, in the spirit of the ROVER algorithm for
combining speech recognition hypotheses. We evaluate the algorithm
for diarization of meeting recordings with multiple microphones, and
find that it consistently reduces diarization error rate over
the average of results from individual channels, and often improves on the single best
channel chosen by an oracle.
\end{abstract}
\section{Introduction}
    \label{sec:intro}

Speaker diarization is the task of segmenting an audio recording in time,
indexing each segment by speaker identity.
In the standard version of the task \cite{TranterReynolds:ieee2006},
the goal is not to identify known speakers, but to 
co-index segments that are attributed to the same speaker;
in other words, the task implies finding speaker boundaries and 
grouping segments that belong to the same speaker
(including determining the number of distinct speakers).
Often diarization is run, in parallel or in sequence, with speech recognition 
with the goal of achieving {\em speaker-attributed speech-to-text}
transcription \cite{FiscusEtAl:nist2007}.

Ensemble classifiers \cite{Rokach:air2010} are a common way
of boosting the performance of machine learning systems, by pooling the
outputs of multiple classifiers.
In speech processing, they have been used extensively whenever 
multiple, separately trained speech recognizers are available, and the goal
is to achieve better performance with little additional integration or modeling
overhead.
The most well-known of these methods in speech processing is ROVER (recognition output voting 
for error reduction) \cite{Fiscus:97}. 
ROVER aligns the outputs of multiple recognizers word-by-word, and then
decides on the most probable word at each position by simple majority or confidence-weighted vote.
Confusion network combination (CNC) is a generalization of this idea that
makes use of multiple word hypotheses (e.g., in lattice or n-best form)
from each recognizer \cite{EvermannWoodland:nist2000,StolckeEtAl:nist2000}.

Given the pervasive use and effectiveness of ensemble methods, it is 
perhaps surprising that so far no ensemble algorithm has been used widely for
diarization.
In this paper we present such an algorithm and apply it to the problem
of combining the diarization output obtained from 
parallel recording channels.
This scenario arises naturally when processing speech captured by 
multiple microphones,
even when the raw signals are combined using beamforming
(because multiple beams can be formed and later combined for improved accuracy,
as described in \cite{Denmark:interspeech2019}).
In a nod to the ROVER algorithm, we call the algorithm DOVER
({\em diarization} output voting for error reduction).
As discussed later, while DOVER is not a variant of ROVER,
a duality can be observed between the two algorithms.

Section~\ref{sec:algo} presents the DOVER algorithm.
Section~\ref{sec:experiment} describes the experiments we ran to test it on
two different datasets involving multi-microphone speech capture.
Section~\ref{sec:conclusion} concludes and points out open problems and future directions.

\section{The Algorithm}
    \label{sec:algo}

\subsection{Motivation and prior work}

The reason that combining diarization outputs in a ROVER-like manner is not 
straightforward is the complex structure of the task:
a diarization system has to perform segmentation (finding speaker boundaries) and 
decisions about identity of speakers across segments.
Where those functions are performed by specialized classifiers inside the diarization algorithm,
ensemble methods could easily be used.
For example, multiple speaker change detectors could vote on a consensus,
or a speaker clustering algorithm could combine multiple acoustic embeddings to evaluate cluster similarity \cite{SunEtAl:icassp2019}.

However, if we are given only the {\em outputs} of multiple diarization processes 
for the same input, or the diarization systems are only available as black boxes,
it is not clear on what part of the output one should ``vote'', and how to combine 
the various hypotheses.

One approach would be to solve diarization as an integer linear programming (ILP) problem
\cite{RouvierMeignier:odyssey2012}.
In ILP-based diarization, a speaker labeling is found that is the best fit to
a collection of local measures of speaker similarity (i.e., the similarity of speech
at times $i$ and $j$ is commensurate with the cost of assigning different speaker labels to $i$ and $j$).
We could translate the different diarization outputs into a set of local similarity 
costs, pool the costs that pertain to the same locations of speech, and then
find a new diarization labeling with ILP.
A similar approach has been used for ensemble segmentation of images \cite{AlushGoldberger:pami2012}.
However, ILP is computationally costly and therefore not widely used in diarization
practice.

The prior method that comes closest to our purpose is a proposal by Tranter \cite{Tranter:icassp2005}, in which pairs of diarization outputs are combined.
The method identifies regions in the audio on which both input diarizations
agree, and passes them through to the output.
Disagreements between the inputs are adjudicated by evaluating speaker identity/nonidentity
according to an external classifier (typically a version of the Bayes information 
criterion, BIC \cite{BIC}).  
Our goal in this work is to reconcile an arbitrary number of diarization outputs,
and to do so using only the outputs 
themselves, without requiring further examination of the acoustic evidence.

\subsection{The DOVER approach}

Our algorithm maps the anonymous speaker
labels from multiple diarization outputs%
\footnote{Without loss of generality, we can assume the labels 
used in the different diarization outputs to be disjoint.}
into a common label space, 
and then performs a simple voting for each region of audio.
A ``region'' for this purpose is a maximal segment delimited by any of the original speaker boundaries, from any of the input segmentations.
The combined (or consensus) labeling is then obtained
by stringing the majority labels for all regions together.

The remaining question is how labels are to be mapped to a common label space.
We do so by using the same criterion as used by the diarization error (DER)
metric itself,
since the goal of the algorithm is to minimize 
the expected mismatch between two diarization label sequences.
Given two diarization outputs using labels $A_1, A_2, \ldots, A_m$ and 
$B_1, B_2, \ldots, B_n$, respectively, an injective mapping
from $\{A_i\}$ to $\{ B_j \}$ is found that minimizes the total time duration
of speaker mismatches, as well as mismatches between speech and nonspeech.%
\footnote{Such an optimal mapping can be found efficiently using a bipartite graph matching algorithm.
In our implementation, we invoke the NIST DER evaluation script \cite{dscore}
as {\tt md-eval.pl -M} to save the mapping to a file.}
Any labels that have no correspondence (e.g., due to differing numbers of speakers) 
are retained.
For more than two diarization outputs, a global mapping is constructed 
incrementally:  after mapping the second output to the labels of the first,
the third output is mapped to the first two.
This is repeated until all diarization outputs are incorporated.
Whenever there is a conflict arising from mapping the $i$th output to each of the prior
$i-1$ outputs, it is resolved in favor of the label pairing sharing the longest 
common duration (overlap in time).

Speech/nonspeech decisions are aggregated by outputting a speaker label if and only if the 
total vote tally for all speaker labels is at least half the total of all inputs,
i.e., the probability of speech is $\geq 0.5$.

\begin{figure}
\newcommand{\DER}{\mbox{\it DER}}
\newcommand{\Label}{\mbox{\it Label}}

\begin{tabbing}
    {\bf Input:} \= A set of $N$ diarization outputs $\{D_i = (L_i, B_i, E_i) \}$, \\
        \> where \= $L_i = \{ l_{ij}, j=1,\ldots,n_i \}$ are speaker labels \\
        \> \> $B_i = \{ b_{ij}, j=1,\ldots,n_i \}$ are segment start times, \\
        \> \> $E_i = \{ e_{ij}, j=1,\ldots,n_i \}$ are segment end times, \\
        \> \> and \= $b_{i,j} < e_{i,j}$ and $e_{i,j} \leq b_{i,j+1}$ \\
        \> \> \> for all $i = 1,\ldots,N$, $j = 1,\ldots,n_i$ \\
        \> \> and $L_j \cap L_k = \emptyset$ for all $j \neq k$ \\
        \> A set of system weights $\{ w_1, \ldots, w_N \}$, with $w_i \geq 0$ \\
    {\bf Algorithm:} \\
    // Label mapping \\
    for \= $i := 2, \ldots, N$ : \\
        \> Label mapping $M := \emptyset$ \\
        \> for \= $k := 1, \ldots, i-1 $ : \\
        \> \> Compute label mapping $M' = \{ (l_j \rightarrow r_j,d_j) \}$ \\
        \> \> \quad \= that minimizes $\DER(D_i,(M'(L_k), B_k, E_k)))$ \\
        \> \>       \> where \= $l_j$ are labels $\in L_i$, \\
        \> \>       \>      \> $r_j$ are labels $\in L_k$, and \\
        \> \>       \>      \> $d_j$ the overlap duration between $l_j$ and $r_j$ \\
        \> \> $M := M \cup M'$ \\
        \> \> Remove from $M$ any mappings $(l \rightarrow r,d)$ \\
        \> \> \quad \= such that there exists a mapping $(l' \rightarrow r,d')$ \\
        \> \>       \> with $l'\neq l, d'\geq d$ \\
        \> $L_i := M(L_i)$ \quad // relabeling \\
    // Label voting \\
    $D^\ast := (\emptyset, \emptyset, \emptyset) $ \quad \= // consensus diarization \\
    $w := \sum_1^N w_i$                     \> // total system weight \\
    for \= all times $t = 0, \ldots, \max(\{e_{i,n_i}\})$ : \\
        \> $l^\ast := \emptyset$ \quad \= // best label \\
        \> $T(l) := 0$ for all labels $l$ \quad // tallies by label \\
        \> for \= $i = 1, \ldots, N$ : \\
        \> \> $l := Label(D_i, t)$ \\
        \> \> $T(l) := T(l) + w_i$ \\
        \> \> if \> $T(l) > T(l^\ast)$ : \\
        \> \>   \> $l^\ast := l$ \\
        \> if \= $T(l^\ast) \geq \frac{1}{2} w $ : \\
        \>  \> // this implies $P(\mbox{speech}) \geq P(\mbox{nonspeech})$ \\
        \>  \> $\Label(D^\ast, t) := l^\ast$ \\
{\bf Output:} \\
    \> Diarization $D^\ast$ \\
\end{tabbing}
    \caption{Pseudo-code for the DOVER algorithm. Notation used:
        $\DER(D_1,D_2)$ is the diarization error function.
        $M(L)$ is the speaker labeling $L$ under label mapping $M$.
        $\Label(D,t)$ is a function that returns the label at time $t$ in diarization $D$, as 
        well as a data structure than can be assigned to, in order to modify the label at the given position in $D$.}
    \label{fig:pseudocode}
\end{figure}

It is straightforward to generalize the algorithm to weighted inputs.
Instead of each input diarization having equal weight (one system, one vote),
the final
voting step adds up the weights attached to the individual systems;
the winning label again is the one with the highest tally.
The weighted-voting version of the algorithm is spelled out in detail in 
Figure~\ref{fig:pseudocode}.

\subsection{An example}

\subsection{Anchoring the label mapping}
    \label{sec:anchoring}

The construction of the global \begin{figure}[tb]
    \centering
    \includegraphics[width=\columnwidth]{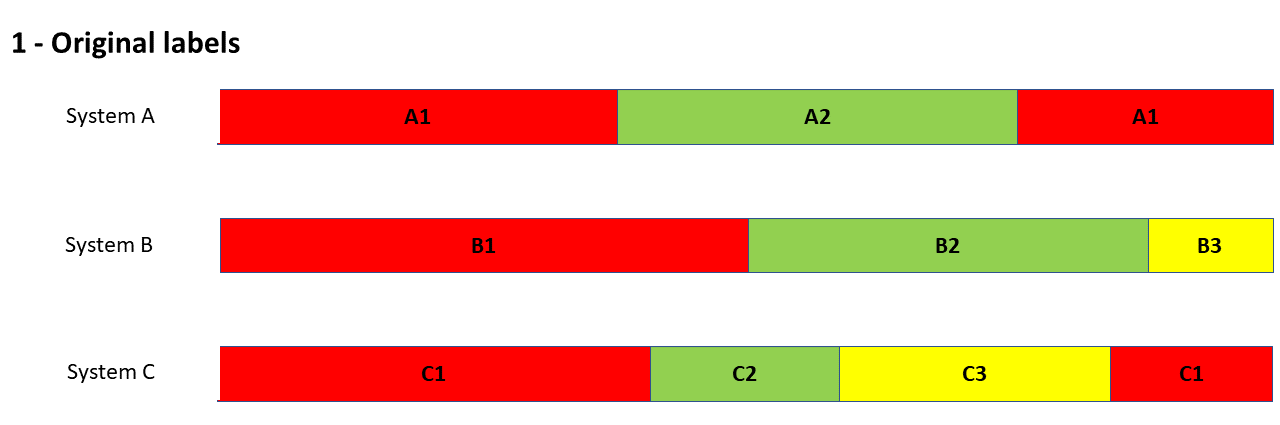}
    \includegraphics[width=\columnwidth]{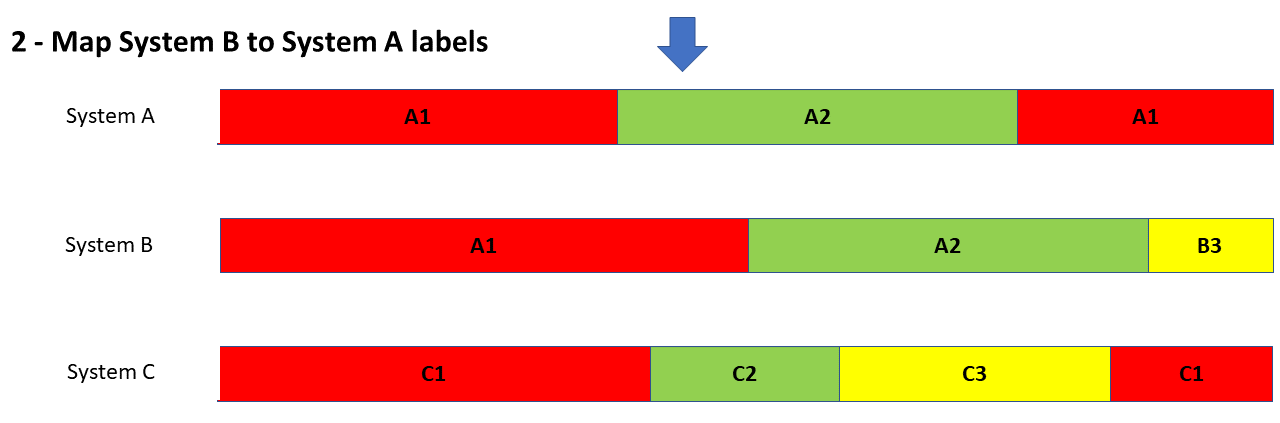}
    \includegraphics[width=\columnwidth]{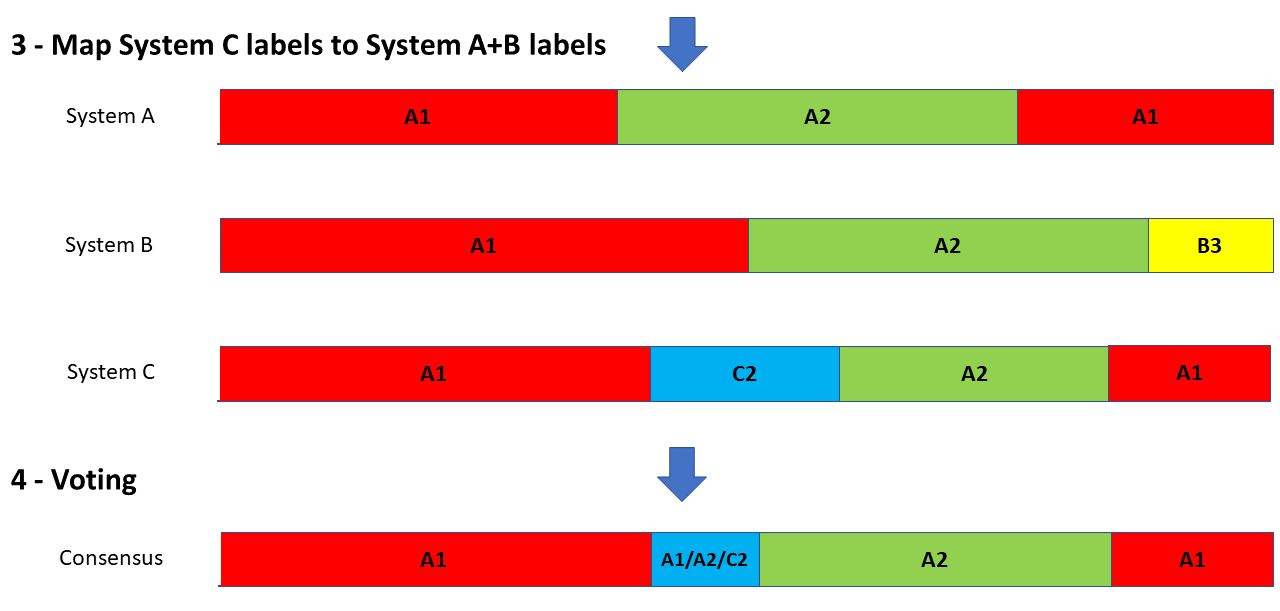}
    \caption{DOVER run on three system outputs (speaker labelings).
            Horizontal extent represents time.
            The original labels are of the form ``B3'', meaning ``speaker 3 from system B''}
    \label{fig:example}
\end{figure}

Figure~\ref{fig:example} shows the workings of the algorithm for three inputs (diarization system outputs) A, B, and C.
For simplicity, non-speech regions are omitted.
Also for simplicity, the inputs are given equal weight.
Step~1 shows the original speaker labelings.
In Step~2 of the algorithm, the labels from System B have been mapped to
labels from System A, using the minimum-diarization-cost criterion.
In Step~3, the output of System C has been mapped to the (already mapped, where applicable) outputs from Systems A and B.
The result is that all three diarization versions now use the same labels where possible,
and in the final step (voting) the consensus labels are determined by taking
the majority label for each segmentation region.

Note that the final output contains one region (shown in blue shading) for which 
no majority label exists, since each of the labels ``A1'', ``A2'' and ``C2''
had only one vote.  In our experiments, we break ties by picking the first label.
Alternatively, a random label could be picked, or the region in question could 
be apportioned equally to the competing labels
(e.g., choosing a temporal ordering that minimizes speaker changes).label mapping is greedy, and dependent on the ordering
of input systems.
(A non-greedy, global optimization of the label mapping for all $N$ inputs would be exponential in the number of inputs $N$.)
The choice of the first input, in particular, could affect the quality of results,
since it anchors the computation of all label mappings.
One strategy is to pick the {\em centroid}, i.e., the diarization hypothesis that
has the smallest aggregate distance (DER) to all the other diarization outputs.
Another, more costly, approach is to run the algorithm $N$ times, once for each input as the anchor.
Then, the $N$ DOVER outputs are themselves combined again (with equal weights) in another run of the algorithm.
For $N$ inputs, this multiplies the overall computation by a factor of $N+1$.

In our experiments we use a variant of the centroid approach:
The input diarization hypotheses are ranked by their average 
DER to all the other hypotheses.
The result is that the centroid comes first, but outlier 
hypotheses also tend to end up at the bottom of the ranking.
We then apply weights to the hypotheses that decay slowly from 1, as a function of rank:
\[
    w_i = \frac{1}{i^{0.1}}
\]
The effect of this is that two lower-ranked hypotheses that 
agree can still override a single higher-ranked hypothesis, but 
ties are broken in favor of the higher-ranked hypothesis.
(If the inputs came with externally supplied ranks, we multiply them
with the rank-based weights.)

\subsection{Duality of DOVER and ROVER}

ROVER and DOVER solve different kinds of tasks:
the former manipulates words labels at discrete positions in a sequence, whereas the latter manipulates anonymous speaker labels positioned on a continuous time axis.%
\footnote{Here we are considering the version of ROVER that does not make use of the time alignment
of the word hypotheses and is based strictly on the edit distance between words strings.}
However, there is an interesting duality between the two algorithms.

In ROVER, the input (word) labels already live in a common name space (the vocabulary) and need to be {\em aligned in time}.
In DOVER, the input (speaker) labels live on a common time axis and need to be
{\em aligned in a common name space} (mapped).
After those two kinds of label alignment are completed, the voting step is similar in the two
algorithms.
Note, also, that the distinction between word sequence and label alignment mirrors the different error metrics.
Word error is mediated by a string alignment that minimizes edit distance.
Diarization error is mediated by a speaker label alignment (i.e., mapping) that minimizes the
sum of speaker and speech/nonspeech error.

\section{Experiments and Results}
    \label{sec:experiment}

\subsection{Data}

We validated the DOVER algorithm on two datasets of meeting recordings with multi microphone channels.
Our focus on this genre of speech is motivated by our overall interest in technology that
can create high-quality speaker-attributed transcripts of multi-person meetings.

The first dataset was drawn from the
NIST 2007 Rich Transcription (RT-07) evaluation \cite{FiscusEtAl:rt07}.
The RT-07 ``conference meeting'' test set consists of 8 meetings from four different recording sites, of varying lengths and with the number of microphones ranging from 3 to 16.
Each meeting has from four to six participants, with 31 distinct speakers in total.
Diarization error is evaluated on a 22-minute speaker-labeled excerpt from each meeting.

The second dataset consists of 5 internal meetings used in Microsoft's ``Project Denmark''
\cite{Denmark:interspeech2019}.
Three of the five meetings were recorded with seven independent consumer devices, followed by automatic synchronization as described in \cite{YoshiokaEtAl:ms-tr2019}.
The other two meetings were recorded with a seven-channel circular microphone array.
The meetings took place in several different rooms and lasted for 30 minutes to one hour each, with three to eleven participants per meeting.
The meetings were neither scripted nor staged;
the participants were familiar with each other and conducted normal work discussions.
The diarization reference labels were derived from time- and speaker-marked transcripts created by professional transcribers based on both close-talking and far-field recordings.

\subsection{Diarization system}

All original diarization outputs for input to DOVER were created with a reimplementation of
the ICSI diarization algorithm \cite{WootersHuijbregts:rt07}.
The algorithm starts with a uniform segmentation of the audio into snippets of equal duration where each segment constitutes its own speaker cluster,
followed by iterative agglomerative clustering and resegmentation.
Distance between speaker clusters is measured by the log likelihood difference between
a single-speaker hypothesis (one Gaussian mixture model) versus the two-speaker hypothesis (two GMMs).
In each iteration, the two most similar speaker clusters are merged, 
followed by a resegmentation of the entire audio stream
by Viterbi alignment to an ergodic HMM over all speaker models.
The merging process stops when a BIC-like criterion \cite{AjmeraEtAL:idiap2002} indicates no further gains in the model likelihood.
When multiple feature streams are used, as described below, the data is modeled by a weighted combination of separate GMMs for each stream.

No attempt is made to detect overlapping speech; therefore all our results have an error
rate floor that corresponds to the proportion of overlapped speech (about 10\% in the Denmark data).

\subsection{Experiments on RT-07 data}

We processed the NIST conference meetings using the weighted delay-and-sum BeamformIt tool \cite{BeamformIt}, 
using $N-1$ audio channels at a time, and resulting in $N$ different audio streams.
This is the same leave-one-out strategy as described in \cite{Stolcke:icassp2011} for
speech recognition.
Furthermore, we rotated the choice of reference channel in these runs to further 
increase diversity among the outputs, as advocated in \cite{YoshiokaEtAl:ms-tr2019}.
We then ran diarization on each of the resulting audio streams, and DOVER on their outputs.
Speech activity was obtained from an HMM-based algorithm that was part of the 
SRI-ICSI meeting recognition system originally used in the RT-07 evaluation \cite{StolckeEtAl:icassp2010}.

Three different feature sets were used in diarization:
\begin{enumerate}
    \item Mel-frequency cepstral coefficients (MFCCs), 19 dimensions, extracted every 10\,ms from the raw waveforms (no beamforming)
    \item MFCCs extracted from the beamformed audio
    \item MFCCs from beamformed audio, augmented with a vector of estimated time-differences-of-arrival (TDOAs) between the different channels, following \cite{AngueraEtAl:ieee2007}
\end{enumerate}

\begin{table}[tb]
    \centering
    \caption{Speaker and diarization error rates on RT-07 meetings.
            All results are macro-averages over the eight test meetings.}
    \label{tab:rt07-results}
    \begin{tabular}{l|c|c|c|c|c}
    \hline
                    &   \multicolumn{3}{c|}{DOVER inputs}       & \multicolumn{2}{c}{DOVER output} \\
    Diarization     &   \multicolumn{3}{c|}{SpkrErr}            & SpkrErr &   DER \\
    inputs          &  max    &  avg. & min &                   &             \\
    \hline
    MFCC (raw audio)& 21.69 & 14.13 & 8.41 & 10.39   & 18.91      \\
    MFCC (BF audio) & 16.80 & 9.43 & 5.48 & 7.04     & 15.58     \\
    MFCC + TDOA     & 12.79 & 5.30 & 2.16 & 2.38     & 10.93     \\ \hline
    \end{tabular}
\end{table}

Table~\ref{tab:rt07-results} shows the outcomes.
The first three result columns give speaker error rates for the individual audio channels.
Note that the ``min'' value is an oracle result, i.e.,
the best that one could do by picking a single channel for diarization.
The last two columns give the speaker error and overall DER for the DOVER-combined 
diarization output.
Note that the difference between speaker error and DER is nearly constant, since all
systems use the same speech activity information.
The missed speech rate is about 3.9\%, while the false alarm rate is 4.6\%.

Looking at the first three columns, 
we observe that the range of error rates is very large (greater than 10\% absolute)
depending on which channel is chosen.
The DOVER-generated diarization has error rates that are closer
to the oracle choice (minimum error) than to the average error,
thereby avoiding the risk of a poor choice of channel.

\subsection{Experiments on Project Denmark data}

For experiments on this dataset, we used byproducts of the Project Denmark meeting transcription system described in \cite{YoshiokaEtAl:ms-tr2019}.
The system aligns the (possibly unsynchronized) audio streams, and then performs
leave-one-out beamforming on 6 out of 7 audio streams, round-robin, resulting in 
7 different new audio streams.%
\footnote{The Denmark beamformer uses a different algorithm than BeamformIt and is based on a neural mask beamformer
adapted from \cite{Boeddeker:icassp2018}; for details see \cite{YoshiokaEtAl:ms-tr2019}.}
For purposes of speaker identification, it then computes 128-dimensional d-vectors
(acoustic speaker embeddings from a neural network trained to perform speaker ID \cite{VarianiEtAl:icassp2014}) at 320\,ms intervals.
The beamformed audio streams are also transcribed by a speech recognition (ASR) system.
Here we use the ASR output only as a speech activity detector (joining words separated by no more than 0.1\,s
of nonspeech, and padding 0.5\,s at the boundaries).

While the Denmark system currently performs speaker identification using enrolled profiles,
we are simulating a scenario where speaker-agnostic diarization is applied instead
(or in addition, e.g, if only a subset of speakers is known to the system).
Since the Denmark audio channels are symmetrical, and no audio channel has privileged status,
we would have to either select one channel for diarization,
or perform diarization on all channels and combine the outputs;
this is where DOVER naturally finds an application.

We ran experiments with three sets of acoustic features, all extracted from the beamformed audio:
\begin{enumerate}
    \item MFCCs, 19 dimensions, extracted every 10\,ms
    \item MFCCs plus the first 30 principal components of the d-vectors (replicated to match the frame-rate of the MFCCs)
    \item MFCCs plus $3 \times 30$ principal components from 3 out of the 7 d-vector streams,
        i.e., a partial feature-level combination of audio streams.
    For channel $i$ the d-vectors were taken from channel $i$ itself, $i-1\pmod{7}$, and $i+1\pmod{7}$.
\end{enumerate}
We also took the outputs of the speaker ID component of the system (from each beamformed audio channel),  treated them as diarization labels, and ran DOVER to see if the algorithm could improve the results.

\begin{table}[tb]
    \centering
    \caption{Speaker and diarization error rates on Denmark meetings.
            All results are macro-averages over the five test meetings.}
    \label{tab:denmark-results}
    \begin{tabular}{l|c|c|c|c|c}
    \hline
                    &   \multicolumn{3}{c|}{DOVER inputs}       & \multicolumn{2}{c}{DOVER output} \\
    Diarization     &   \multicolumn{3}{c|}{SpkrErr}            & SpkrErr &   DER \\
    inputs          &  max    &  avg. & min &                   &                   \\
    \hline
    MFCC                & 34.56 & 24.23 & 15.56 & 15.00     & 26.98     \\
    \  + d-vector     & 13.94 & 11.06 & 8.82  & 8.70      & 20.65     \\
    \  + 3 d-vectors  & 11.38 & 6.07  & 3.00  & 3.10      & 14.97     \\
    \hline
    Speaker ID        & 2.18  & 1.86  & 1.42  & 1.20      & 13.06     \\ \hline
    \end{tabular}
\end{table}

Table~\ref{tab:denmark-results} shows the results for diarization based on the three feature set,
as well as based on speaker ID, using the same format as for the RT-07 results.
Here, too, the difference between speaker error and DER is nearly constant, since all
systems use the same speech activity information derived from the speech recognizer.
The DER thus includes about 0.6\% false alarms and 11.3\% miss rate (of which 10.0\% are due to overlapped 
speech, which we do not attempt to detect).

The most important observation is that the DOVER output has a speaker error rate that is very close to,
and for the most part slightly lower than, the best (oracle) choice of channel.
As for the RT-07 data, the DOVER output is consistently much better than the channel average.
Also, the max values show that there is still ample opportunity for very poor choices of a single channel; DOVER removes the need to make that choice.

The last row of results shows that even when the diarization on individual channels is very accurate (due to the availability of speaker models), DOVER can still give a substantial relative error reduction, surpassing the best channel's performance.

\section{Conclusions and Outlook}
    \label{sec:conclusion}
    
We have presented a weighted voting algorithm for combining the outputs from several diarization systems
over a shared input.
The DOVER algorithm first uses a DER-minimizing criterion to map all speaker labels to a common name space,
and then performs majority voting at each time instant (including on whether there is speech or not).
The proposed method naturally lends itself to unifying diarization outputs obtained from
parallel audio channels, e.g., as they arise from meeting capture with multiple microphones or devices.
We tested the algorithm on a NIST conference meeting evaluation set, as well as on internal meetings,
using diarization by agglomerative clustering combined with a variety of feature streams.
We find that the DOVER output consistently beats the averages of the input channels, and can be 
very close or improving on the oracle error rate obtained by picking the single best channel for a given
meeting.

Some interesting open issues remain.
As mentioned, we currently do not attempt to diarize overlapping speech. Once such 
a capability is available, the DOVER algorithm will have to be modified to handle simultaneous speakers.
Another issue is that current diarization systems only output their single best guesses at the speaker labeling.
In analogy to confusion network combination, we may want to consider diarization algorithms that
produce multiple weighted hypotheses, which are then in turn combined across all systems.
A modified DOVER could be used both to generate the ``speaker confusion networks'' from individual 
diarization systems, and to combine them.

\section*{\hspace{0.3\columnwidth}Acknowledgments}
We thank our colleagues for help with the Denmark system and data collection,
Xavi Anguera for answering questions regarding BeamformIt,
and ICSI for assistance with the RT-07 data.

\bibliographystyle{ieee-shortnames}
\bibliography{refs}

\end{document}